\documentclass[10pt,twocolumn,letterpaper]{article}

\usepackage[pagenumbers]{wacv}  % To force page numbers, e.g. for an arXiv version

\usepackage{graphicx}
\usepackage{amsmath}
\usepackage{amssymb}
\usepackage{booktabs}
\usepackage{multirow}
\usepackage{makecell}
\usepackage{upgreek}

\usepackage[pagebackref,breaklinks,colorlinks]{hyperref}

\usepackage[capitalize]{cleveref}
\crefname{section}{Sec.}{Secs.}
\Crefname{section}{Section}{Sections}
\Crefname{table}{Table}{Tables}
\crefname{table}{Tab.}{Tabs.}

\def\wacvPaperID{*****} % *** Enter the WACV Paper ID here
\def\confName{WACV}
\def\confYear{2024}

\begin{document}

%%%%%%%%% TITLE
\title{Analyzing the Optimization Conflict Between\\Image- and Text-Based Person Re-Identification}

\author{
Karina Kvanchiani$^{1*}$ \\
{\tt\small karinakvanciani@gmail.com}
\and
Timur Mamedov$^{1,2*}$ \\
{\tt\small me@timmzak.com}
\and
$^1$Tevian, Moscow, Russia \quad $^2$Lomonosov Moscow State University, Russia}

\maketitle
\renewcommand{\thefootnote}{}
\footnotetext{$^*$Both authors contributed equally.}
\renewcommand{\thefootnote}{\arabic{footnote}}

%%%%%%%%% ABSTRACT
\begin{abstract}
   Unifying image-based (I2I) and text-based (T2I) person re-identification (ReID) within a single model is complicated by a fundamental conflict between their training objectives. While I2I ReID focuses on identity-level invariance across images of the same person captured by multiple cameras, T2I ReID is driven by instance-specific textual descriptions tied to unique visual traits. When both objectives are optimized on a shared vision encoder, they impose opposing requirements on the same embeddings — I2I contracts intra-identity distances while T2I requires them to persist — so improving one task's representation actively degrades the other's. This paper analyzes this optimization conflict and shows empirically that the two objectives are not jointly satisfiable by a single unconstrained encoder. Motivated by this finding, we investigate the conflict through a controlled two-stage pipeline: a shared vision encoder is first trained for I2I invariance, then frozen while only the text encoder is aligned to the fixed visual space, preventing the T2I objective from unwinding the identity-level structure already learned. We conduct extensive experiments across multiple configurations, varying domain-mixing procedures, learning strategies, and task objectives. We observe that I2I pre-training improves subsequent T2I performance, while textual supervision during this pre-training further improves T2I performance, although the vision encoder is frozen during subsequent cross-modal training. We believe these insights provide a meaningful step toward unified ReID systems that reconcile identity-level and instance-level supervision within a single representation.
\end{abstract}

%%%%%%%%% BODY TEXT
\section{Introduction}
\label{sec:introduction}
Person Re-Identification~(ReID) is a retrieval problem that aims to match a query and corresponding images of the same individual. The search can be conducted in various ways depending on the query modality. Traditional Image-to-Image~(I2I, image-based) ReID focuses on learning visual representations that are robust to variations in pose, illumination, and background across multiple camera views. In contrast, more flexible and human-friendly Text-to-Image~(T2I, text-based) ReID is based on cross-modal matching between text queries and searched images.

\begin{figure}[t]
  \centering
  \includegraphics[width=\linewidth]{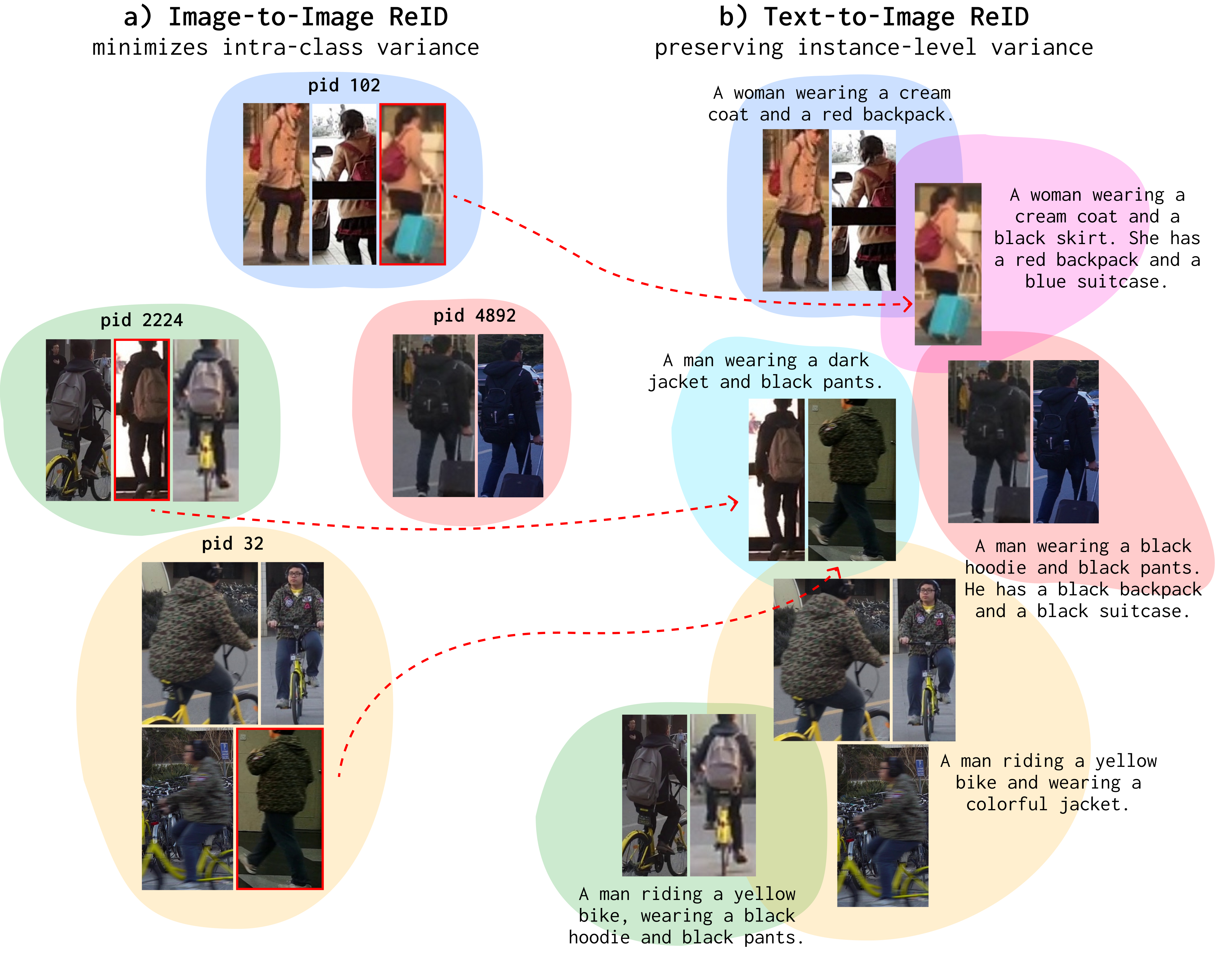}
  \caption{Conflicting supervision signals in I2I and T2I ReID. a)~In I2I ReID, all images of the same person are treated as equivalent positives, pulling every instance toward a single identity centroid and minimizing intra-class variance. b) In T2I ReID, each caption is grounded in a specific visual observation, so instances of the same identity must remain separable to be matched to their respective descriptions.}
  \label{fig:main3}
\end{figure}

\begin{figure}[t]
  \centering
  \includegraphics[width=\linewidth]{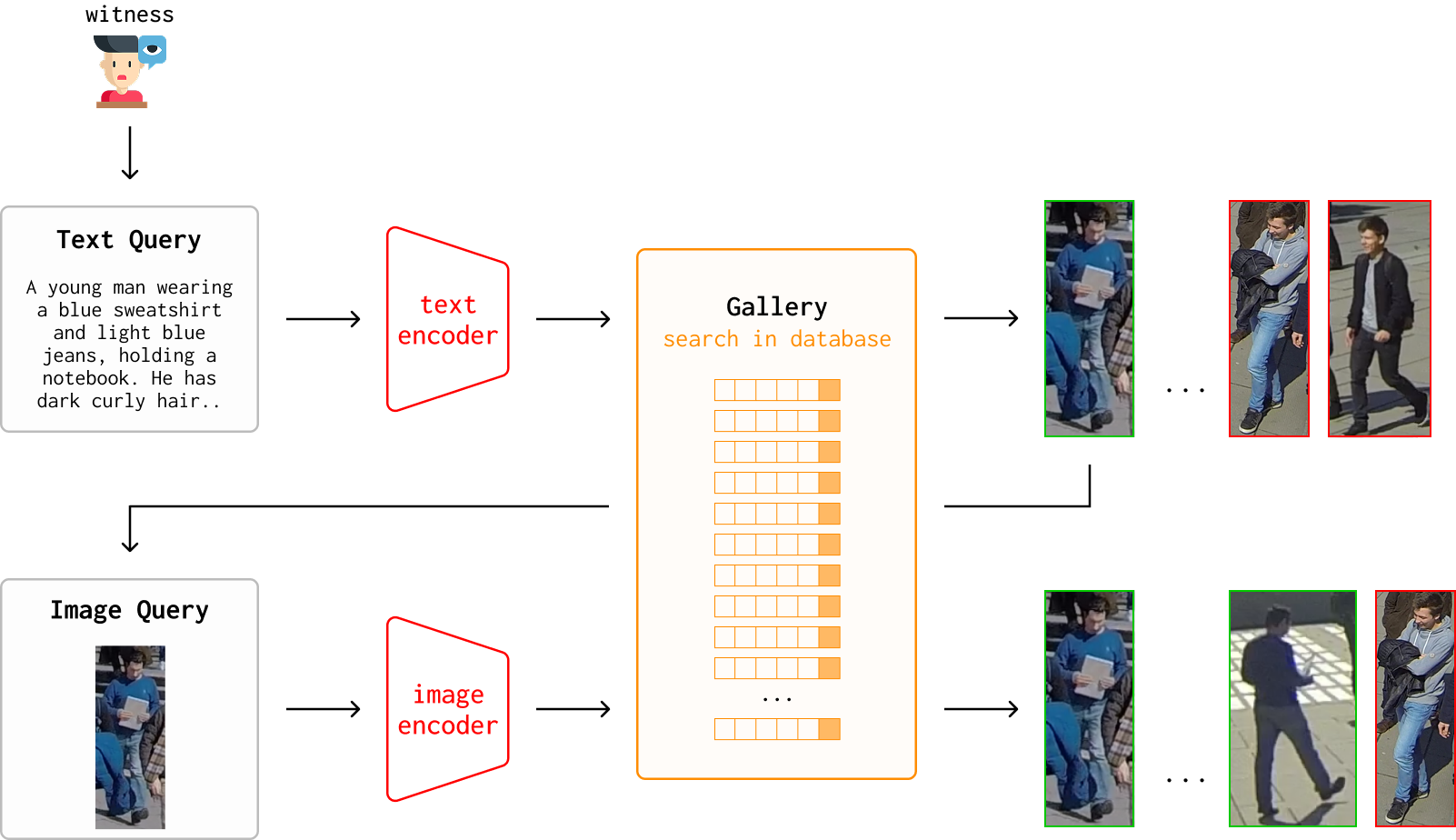}
  \caption{One practical scenario of a unified ReID system. The gallery includes image embeddings generated by an image encoder, trained jointly with the text one.}
  \label{fig:demo}
\end{figure}

Despite their apparent similarity, I2I and T2I ReID rely on fundamentally different supervision signals. Image-based ReID treats all images of the same person as equivalent positives and enforces invariance to intra-identity variations~(Fig.~\ref{fig:main3}a). Additionally, I2I ReID is usually complicated by multi-camera data, which involves images of the same identity captured across different camera views, introducing variations in pose, illumination, and background. This setting is essential for learning identity-invariant representations. In contrast, T2I ReID is instance-level, as textual descriptions are aligned with a specific visual observation rather than the underlying identity. As a result, attributes mentioned in the text may not hold across different appearances of the same person. For example, a person described as “riding a yellow bike” in one image may appear without the bike in another one~(Fig.~\ref{fig:main3}b). While both images represent the same identity in I2I ReID, they show different instances from the T2I perspective. Moreover, a caption for one identity may be semantically close to images of another identity. Aligning these images with the caption can unintentionally pull different identities closer together, opposing the goal of I2I ReID.

This distinction also shapes the available training data. T2I ReID typically relies on single-camera datasets, which are too simplistic for learning identity-invariant representations, while multi-camera data, essential for I2I ReID, generally lacks textual annotations. As a result, the two tasks are usually trained on distinct data sources.

Although there is extensive research on each task separately, it is beneficial to support both image- and text-based queries within a single ReID model in the surveillance system, as it avoids deploying and maintaining separate models for different query modalities. Besides, one of the most practical scenarios arises when no image query is available, and the user starts the search with a natural-language description of a person, expecting the system to provide an initial set of visual candidates (Fig.~\ref{fig:demo}). Then, the most suitable candidate can be utilized as an image query for subsequent I2I retrieval, which is more precise in terms of identity-level representation.

\begin{figure}[t]
  \centering
  \includegraphics[width=0.95\linewidth]{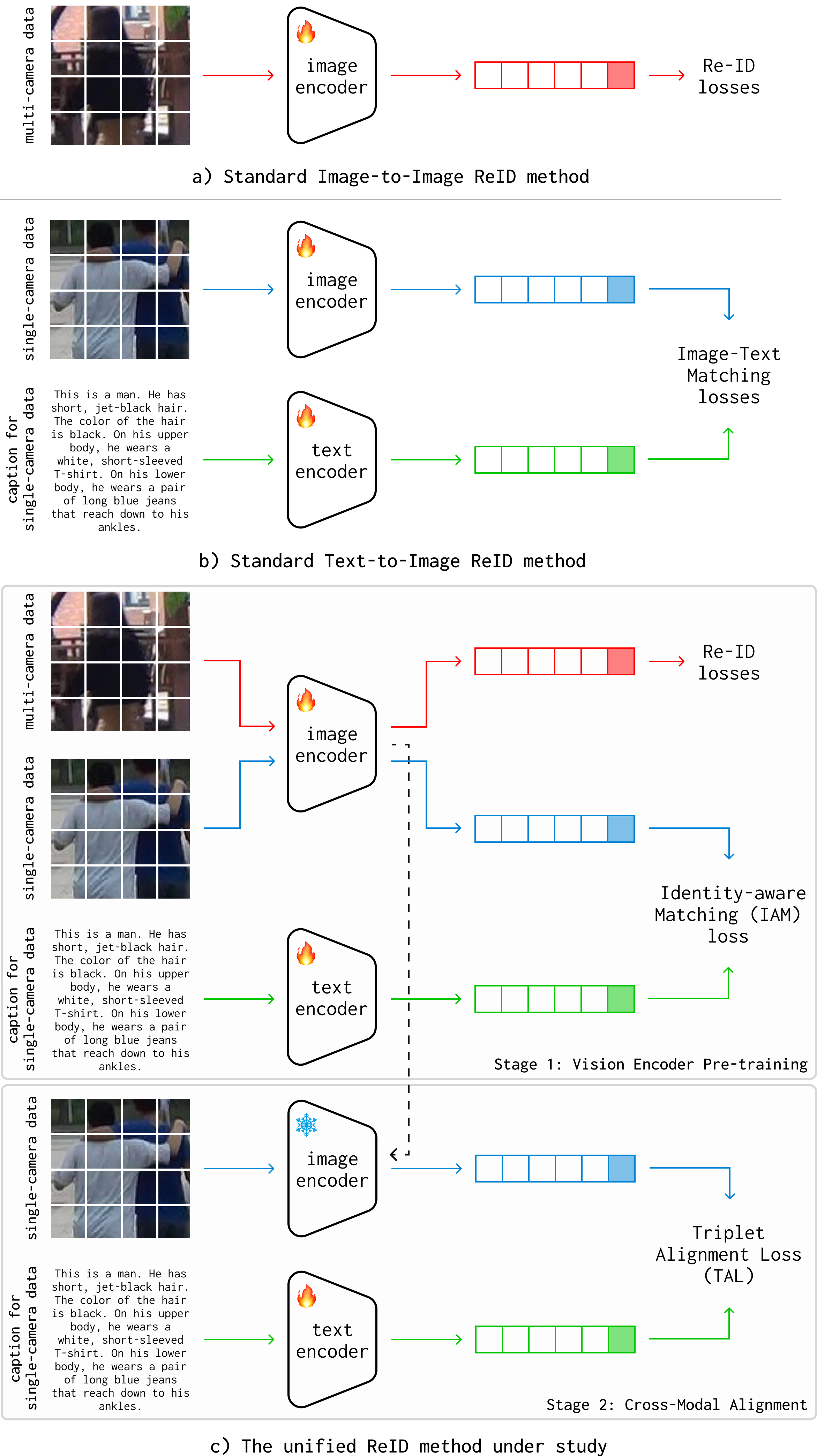}
  \caption{The difference between the standard I2I-only and T2I-only methods and the unified method under study. a) The vision encoder receives multi-camera data and is trained with ReID losses. b) The model obtains single-camera images annotated with textual descriptions to align embeddings of different modalities through image-text matching losses. c) The vision encoder, pre-trained on the combination of multi-camera and single-camera data with ReID and image-text matching losses, is frozen, and the model learns to align text embeddings to its representation space.}
  \label{fig:main}
\end{figure}

Typically, image-based ReID relies on a vision encoder~(Fig.~\ref{fig:main}a), while text-based ReID additionally requires a text encoder to project language descriptions into a shared embedding space (Fig.~\ref{fig:main}b). Since both tasks rely on visual representations, it is natural to research whether a single, shared vision encoder could support both I2I and T2I ReID. However, as discussed above, the two tasks rely on fundamentally different supervision signals, so naively training one encoder for both raises the question of whether their objectives can coexist without mutual degradation.

Motivated by these observations, we revisit the consideration of the ReID frameworks that jointly support both I2I and T2I retrieval within a single model. Instruct-ReID~(IRM)~\cite{irm} and HPL~\cite{hpl} approach this through architectural means --- instruction following and task-aware prompt learning, respectively --- but neither analyzes whether the I2I and T2I ReID objectives interfere, leaving the source of their gains unexplained. In contrast, we provide a comprehensive analysis of a conflict that arises when jointly optimizing image- and text-based ReID objectives. The contributions of this work can be summarized as follows:

\begin{itemize}
    \item We study the fundamental differences between I2I and T2I ReID,  analyze the emerging optimization conflict, and show empirically that the two objectives are not interchangeable. Understanding the problem is key to achieving competitive performance while avoiding additional architectural complexity.
    \item We study the conflict through
    a controlled two-stage pipeline built on a single vision encoder, enabling unified retrieval (Fig.~\ref{fig:main}c), and analyze
    the effects of encoder freezing.     \item Through a series of controlled ablations, we examine such factors as the interchangeability of image-text matching losses across stages, the influence of textual supervision during vision encoder pre-training, and the impact of domain overlap between the two stages.
\end{itemize}

\section{Related Work}
\subsection{Image-based ReID}
I2I ReID learns visual representations invariant to intra-identity variation in pose, illumination, and viewpoint, so that all images of the same person collapse toward a common region of the embedding space. TransReID \cite{transreid} tackled this by introducing a pure-transformer architecture with side-information embeddings and a jigsaw patch module to capture robust part-level cues. To reduce the reliance on manual identity labels, self-supervised and vision-language pre-training strategies have been explored. PASS \cite{pass} performed part-aware self-supervised pre-training, which produces fine-grained local features transferable to the downstream ReID task. At the same time, CLIP-ReID \cite{clipreid} adapted a vision-language model to image-based ReID using learnable text prompts that enriched vision embeddings with semantic information. More recent work targeted generalization across heterogeneous data sources. ReMix \cite{remix} trained a generalized model on a mixture of multi-camera and single-camera ReID datasets, while DynaMix \cite{dynamix} improved its cross-dataset generalization through dynamic relabeling and mixed-data sampling. ReText \cite{ReText} extended this line by incorporating textual supervision during vision encoder training. Unlike CLIP-ReID, which optimizes learnable text prompts, ReText leverages natural-language descriptions of single-camera data, exploiting their full semantic content. In this work, we adopt ReText's vision encoder because it saturates the visual space with identity-level semantics while remaining open to textual cues, providing a strong anchor for subsequent cross-modal alignment.

\subsection{Text-based ReID}
T2I ReID retrieves person images from natural-language descriptions and is inherently instance-level, i.e., each caption is grounded in one specific visual observation, so attributes it mentions need not persist across other images of the same identity. A central challenge is therefore fine-grained cross-modal alignment. IRRA~\cite{irra} tackled this by learning implicit relation reasoning between local textual and visual tokens and aligning modalities through a similarity-distribution matching objective. Building on the observation that real captions are often imperfectly aligned, RDE~\cite{RDE} addressed the resulting noisy-correspondence problem with a contrastive, CLIP-like design that relaxes the ranking of hard negatives via its Triplet Alignment Loss~(TAL). In a similar spirit of handling unreliable supervision, RaSa~\cite{rasa} introduced relation- and sensitivity-aware representation learning to mitigate weak positive pairs. A complementary direction instead strengthens the alignment signal itself. MARS~\cite{mars} placed additional emphasis on visual attributes to reinforce text--image correspondence, and CFAM~\cite{cfam} refined attribute-level matching for benchmarks that demand detailed discrimination. The granularity of descriptions is a related challenge. VISA~\cite{visa} addressed granularity mismatches between textual queries and visual content in text-visual retrieval, while LLaVA-ReID~\cite{llavareid} obtained fine-grained person descriptions through interactive questioning. Despite their differing emphases, these methods consistently show that instance-level, triplet-style objectives best capture the fine textual distinctions that T2I retrieval depends on. The second stage of the pipeline under study is built on the RDE framework and its Triplet Alignment Loss for cross-modal alignment.

\subsection{Unified ReID}
A single model that serves both image and text queries avoids deploying and maintaining separate systems, motivating recent unified ReID frameworks. Instruct-ReID~(IRM) \cite{irm} reformulated ReID as an instruction-following problem and retrieved same-identity images from query images and multimodal instructions. However, the impact of instruction tuning on identity preservation and cross-modal alignment is not analyzed, nor are potential conflicts between I2I and T2I ReID tasks. Another method, HPL~\cite{hpl}, employed task-aware prompt learning to align visual and textual representations, reporting consistent gains across both retrieval settings. The source of these gains, however, is not examined — the method does not analyze how identity-level and instance-level supervision interact when the two tasks share a representation. Unlike these approaches, which pursue an architectural solution, we treat the unified setting as an optimization problem. We show that identity-level (I2I) and instance-level (T2I) objectives actively conflict when jointly optimized on a shared vision encoder, and we study a decoupled two-stage pipeline that avoids this conflict: the encoder is first pre-trained for I2I and then frozen while only the text encoder is aligned to its fixed space. This preserves I2I discriminability, remains competitive with HPL, and consistently surpasses IRM, without adding architectural complexity.

\subsection{Instance- vs semantic-level objectives}
Standard contrastive training couples both modalities under instance-level objectives, leading to a conflict between instance discrimination and the semantic-level grouping of multiple relevant items per query. While most modern methods treat each image-text pair independently, some works show that instance-level objectives fail to capture relationships among semantically similar samples. For example, ~\cite{ir1} introduced support-set regularization to explicitly model multiple related samples, while ~\cite{ir2} used neighborhood-aware re-ranking to recover semantic structure in the embedding space. Two other works~\cite{ir4,ir5} mitigated this issue by relaxing instance-level objectives. While ~\cite{ir4} showed that standard contrastive loss neglects intra-modality structure,~\cite{ir5} presented softer alignment methods for better semantic relationships. Additionally, ~\cite{ir6} preserved the learned semantic structure by freezing the pre-trained backbone during cross-modal alignment.

A structurally identical tension has been identified in self-supervised representation learning. Standard instance discrimination objectives treat every image as its own class, pulling together only its augmented views while pushing apart all other samples, including other instances that share the same semantic class. SCL~\cite{scl} identified this as an inherent contradiction between instance-level discrimination, which pulls same-cluster samples apart, and cluster-level grouping, which requires pulling them closer, and showed that this conflict is responsible for the class ambiguities observed when contrastive objectives are used for deep clustering. ReCo~\cite{reco} similarly showed that vanilla contrastive self-supervised learning inevitably pushes away samples from the same semantic category as negatives, degrading the fine semantic structure of the learned representation, and addressed this by explicitly modeling instance relations rather than treating all non-augmented samples as equally negative.

Our work identifies the same structural conflict within person ReID, where identity-level I2I supervision serves as cluster-level grouping, pulling same-identity images together, while instance-level T2I supervision serves as instance discrimination, requiring those same images to remain separable. While \cite{scl,reco} addressed optimization conflicts by proposing mitigation strategies, they did not investigate the conflict itself. We instead explicitly study the optimization conflict between image-based and text-based retrieval objectives.

\section{Problem Formulation}
\label{sec:problem}
We first formalize the I2I and T2I retrieval settings and the optimization conflict that arises when a single visual representation needs to serve both.

\subsection{Notation}
\label{sec:notation}
Let $\mathcal{X}$ be the image space and $\mathcal{T}$ the text space. A shared vision encoder $f_\theta:\mathcal{X}\!\to\!\mathbb{R}^{d}$ maps an image $x$ to an $\ell_2$-normalized embedding $\mathbf{v}=f_\theta(x)/\lVert f_\theta(x)\rVert$, and a text encoder $g_\phi:\mathcal{T}\!\to\!\mathbb{R}^{d}$ maps a caption $t$ to $\mathbf{u}=g_\phi(t)/\lVert g_\phi(t)\rVert$ in the same space. Retrieval ranks candidates by cosine similarity $s(a,b)=\langle a,b\rangle$.

Each image $x$ is associated with an identity label $y\in\mathcal{Y}$ and, in the multi-camera setting, a camera label $c\in\mathcal{C}$. We write $x_i\!\sim\!x_j$ when $y_i=y_j$ and $x_i\!\not\sim\!x_j$ otherwise. For an anchor $x_i$ we denote its identity-positive set $\mathcal{P}^{\text{id}}_i=\{x_j : y_j=y_i,\, j\neq i\}$ and its negative set $\mathcal{N}_i=\{x_j : y_j\neq y_i\}$. In the text-based setting, each caption $t_i$ describes one specific image, which we denote by the same index, $x_i$. All other images, including those of the same person, are denoted $x_j$ with $j \neq i$.

\subsection{I2I ReID as an Identity-level Supervision}
\label{sec:i2i_def}
I2I ReID treats all images of one identity as equivalent positives and seeks a representation that is invariant to intra-identity variation in pose, illumination, viewpoint, and camera. Formally, the ideal I2I space satisfies, for every anchor $x_i$,
\begin{equation}
s(\mathbf{v}_i,\mathbf{v}_j) > s(\mathbf{v}_i,\mathbf{v}_k), \quad \forall\, x_j\in\mathcal{P}^{\text{id}}_i,\ \forall\, x_k\in\mathcal{N}_i,
\label{eq:i2i_rank}
\end{equation}
independently of the camera labels $c_i,c_j$. Equivalently, the objective drives every instance of an identity toward a single identity centroid $\boldsymbol{\mu}_{y}= \mathbb{E}_{x:y_x=y}[\mathbf{v}_x]$, minimizing the intra-class variance
\begin{equation}
\sigma^2_{\text{intra}}(y)=\mathbb{E}_{i:\,y_i=y}
\big[\,1-s(\mathbf{v}_i,\boldsymbol{\mu}_{y})\,\big].
\label{eq:intra}
\end{equation}
The identity is thus the atomic unit, i.e., I2I supervision is invariant to which particular instance of $y$ is observed, and collapsing the instances of an identity is the desired behavior.

\subsection{T2I ReID as an Instance-level Supervision}
\label{sec:t2i_def}
T2I ReID matches a caption to the images of the person it describes. Unlike I2I, however, the supervision is instance-grounded, where a caption $t_i$ is written for one specific observation $x_i$, so the attributes it mentions (e.g., a carried umbrella) need not hold across other images of the same identity. Across different identities, the requirement is the familiar ranking condition,
\begin{equation}
s(\mathbf{u}_i,\mathbf{v}_i) > s(\mathbf{u}_i,\mathbf{v}_k), \quad \forall\, k \text{ with } x_i\!\not\sim\!x_k,
\label{eq:t2i_rank}
\end{equation}
but the informative case lies within an identity. Consider another image $x_j$ of the same person ($y_j = y_i$, $j \neq i$) in which the described attribute is absent. For the caption to remain grounded in the observation it actually describes, the similarity gap
\begin{equation}
\Delta_i = s(\mathbf{u}_i,\mathbf{v}_i) - s(\mathbf{u}_i,\mathbf{v}_j)
\label{eq:delta}
\end{equation}
must be strictly positive. The individual observation, not the identity, is thus the atomic unit of T2I supervision.

Two clarifications are in order. First, captions of other images of the same person are not true negatives but weak positives --- they share identity-stable attributes such as clothing or build, while differing in instance-specific ones such as a carried object or a pose. Matching objectives used in T2I ReID reflect this by assigning such pairs soft rather than zero weight. Second, at inference, the ground truth for a text query is the full set of images of the corresponding identity, so the evaluation target is identity-level even though the supervision is instance-grounded. The gap $\Delta_i$ should therefore be read not as a requirement on the final ranking, but as a diagnostic of whether the shared space still encodes instance-level information. When $\Delta_i \to 0$, the described attribute carries no discriminative signal, and the fine-grained grounding on which cross-modal alignment depends is lost --- regardless of how the retrieved set is subsequently scored.

\subsection{The Optimization Conflict}
\label{sec:conflict_def}
In joint training of a model addressing both I2I and T2I ReID tasks, the two objectives act on the same visual embeddings $\{\mathbf{v}_i\}$ but impose opposite requirements on the geometry of an identity. I2I ReID minimizes intra-identity variance~\eqref{eq:intra}, collapsing all images of a person toward the centroid $\boldsymbol{\mu}_{y}$. T2I ReID, in contrast, requires those same images to remain instance-separable~\eqref{eq:delta}, so that each caption stays grounded in the observation it describes. Concretely, as I2I ReID supervision drives $\sigma^2_{\text{intra}}(y)\!\to\!0$, same-identity embeddings converge, $\mathbf{v}_i\!\approx\!\mathbf{v}_j$ for $y_j = y_i$, forcing
\begin{equation}
\Delta_i = s(\mathbf{u}_i,\mathbf{v}_i) - s(\mathbf{u}_i,\mathbf{v}_j)
\;\longrightarrow\; 0 ,
\label{eq:conflict}
\end{equation}
so that the described attribute no longer carries a discriminative signal and the fine-grained grounding on which cross-modal alignment depends is erased. The two objectives are therefore not jointly satisfiable by a single unconstrained encoder, i.e., strengthening identity-level invariance erodes the instance-level separability required by text, and vice versa.

\section{Methodology}
\label{sec:method}
This work aims at a thorough analysis of the optimization conflict rather than proposing an architecture. We study the problem from different perspectives by adapting existing approaches to extract actionable insights. We start our research with the T2I ReID method, as the architecture of such approaches inherently encompasses the vision encoder required for I2I ReID. We choose RDE~\cite{RDE}, since it is a strong and readily adaptable baseline for T2I ReID. It follows a CLIP-like paradigm to jointly train vision and text encoders using a contrastive objective, aligning paired image-text samples in a shared embedding space. However, the T2I ReID method is insufficient for an effective unified model~(Sec.~\ref{sec:rde_i2i}). So, we propose training the vision encoder separately on I2I data using the image-based ReID approach. Then, saturated visual representations with identity-level semantics are reused to train a text encoder with projection in the RDE setup.

To saturate visual representations with multi-camera semantics, we utilize the recently introduced I2I framework named ReText~\cite{ReText}. In addition to achieving strong performance in a domain generalization setup, ReText enhances I2I ReID by incorporating textual supervision during the vision encoder training. Unlike CLIP-style training, ReText does not enforce strict one-to-one matching between images and text. Instead, textual descriptions serve as auxiliary supervision to enrich visual representations with semantic information, while focusing on grouping images by identity. This allows the model to incorporate semantic cues without disrupting identity invariance. Despite the presence of a text encoder, our experiments show that ReText is ineffective for addressing the T2I ReID problem (Tab.~\ref{tab:sota}).

Taking all of the above into account, we divide the studied pipeline into two stages. In the first stage, we pre-train the vision encoder via ReText, while in the second, we adopt it as the backbone within the RDE framework~(Fig.~\ref{fig:main}c). To preserve the discriminative power of image-based ReID and avoid degradation in I2I ReID performance, the pre-trained vision encoder is frozen during the cross-modal training stage. This design allows us to leverage text-enriched visual representations while maintaining strong identity-level invariance required for I2I ReID. Less restrictive alternatives to full freezing are compared in Section~\ref{sec:freeze}.

\subsection{Stage 1: Vision Encoder Pre-training}
The ReText training strategy can be divided into two parts based on data type. A multi-camera data branch is based on images of the same identity captured across different camera views, introducing variations in pose, illumination, and background to optimize I2I ReID objectives. This setting is essential for learning identity-invariant representations. In contrast, the single-camera data branch is less demanding about various views, but requires text information for each image to incorporate it into the vision encoder.

\vspace{2pt}\noindent\textbf{Multi-camera I2I ReID Branch.}
ReText optimizes the I2I ReID task on multi-camera data to enforce cross-view invariance via four objectives:
\begin{equation}
\mathcal{L}_{ReID} = \mathcal{L}_{ins} + \mathcal{L}_{aug} + \mathcal{L}_{cen}
+ 0.5\,\mathcal{L}_{cc}.
\label{eq1}
\end{equation}
Let $\mathbf{v}_i$ denote the $\ell_2$-normalized embedding of image $x_i$ produced by the encoder, and $\tilde{\mathbf{v}}_i$ the corresponding embedding produced by the momentum encoder, whose weights are an exponential moving average of the encoder's for stable training.

The Instance Loss enforces separation between identities by pulling all positives of an anchor together and pushing negatives apart:
\begin{equation}
\mathcal{L}_{ins} = \frac{1}{B}\sum_{i=1}^{B}\mathcal{L}^{i}_{ins},
\label{eq21}
\end{equation}
\begin{equation}
\mathcal{L}^{i}_{ins} = \frac{-1}{K}\sum_{j:\,y_j=y_i}
\log\frac{\exp\!\big(s(\mathbf{v}_i,\tilde{\mathbf{v}}_j)/\tau_{ins}\big)}
{\sum_{k=1}^{N+1}\exp\!\big(s(\mathbf{v}_i,\tilde{\mathbf{v}}_k)/\tau_{ins}\big)},
\label{eq22}
\end{equation}
where $K$ is the number of images per identity in the mini-batch, $N$ is the number of negatives for the anchor $i$, and $\tau_{ins}$ is a temperature controlling the sharpness of the similarity distribution. The sum in the numerator runs over all images sharing the anchor's identity, so every positive contributes rather than only the hardest one.

The Augmentation Loss preserves identity consistency under data augmentations:
\begin{equation}
\mathcal{L}_{aug} = \mathbb{E}\!\left[-\log\frac{\exp\!\big(s(\mathbf{v}^{aug}_i,\tilde{\mathbf{v}}_i)/\tau_{aug}\big)}{\sum_{j=1}^{N+1}\exp\!\big(s(\mathbf{v}^{aug}_i,\tilde{\mathbf{v}}_j)/\tau_{aug}\big)} \right],
\label{eq3}
\end{equation}
where $\mathbf{v}^{aug}_i$ is the embedding of an augmented view of $x_i$ and $\tilde{\mathbf{v}}_i$ is the momentum embedding of its original view, which serves as the positive target. $\tau_{aug}$ is the corresponding temperature. The loss thus keeps the augmented and original views of the same image aligned while separating them from other identities.

The Centroids Loss encourages compact identity clusters around class centroids:
\begin{equation}
\mathcal{L}_{cen} = \frac{1}{B}\sum_{i=1}^{B}\mathcal{L}^{i}_{cen},
\label{eq41}
\end{equation}
\begin{equation}
\mathcal{L}_{cen}^{i} = -\log \frac{\exp\!\big(s(\mathbf{v}_i,\boldsymbol{\mu}_{y_i})/\tau_{cen}\big)} {\sum_{y\in\mathcal{Y}}\exp\!\big(s(\mathbf{v}_i,\boldsymbol{\mu}_{y})/\tau_{cen}\big)}.
\label{eq42}
\end{equation}
where $\boldsymbol{\mu}_{y}$ is the centroid of the momentum embeddings of all images with identity $y$, $\mathcal{Y}$ is the set of identities in the mini-batch, and $\tau_{cen}$ is a temperature. Unlike~\eqref{eq21}, which operates on individual pairs, this objective compares each sample against identity prototypes, directly minimizing the intra-class variance defined in~\eqref{eq:intra}.

Finally, the Camera Centroids Loss $\mathcal{L}_{cc}$~\cite{cc_loss} extends this idea to the camera dimension. Instead of a single centroid per identity, it maintains camera-specific centroids and pulls each sample toward the centroids of its own identity computed from other cameras, while pushing it away from the centroids of different identities. This explicitly reduces the intra-class variance induced by stylistic differences between camera views.

\vspace{2pt}\noindent\textbf{Single-camera T2I ReID Branch.} In parallel, single-camera datasets annotated with natural language descriptions are incorporated to provide instance-level textual supervision. The loss functions presented in ReText encourage robust image-text alignment, preserve the structural consistency of visual representations, and ground the vision encoder in semantic cues from language. Throughout this branch, $B$ denotes the number of image-text pairs in the mini-batch, $\mathbf{v}_i$ the $\ell_2$-normalized embedding of image $x_i$, and $\mathbf{u}_i$ the $\ell_2$-normalized embedding of its caption $t_i$.

First, the Identity-aware Matching Loss (IAM) relaxes strict one-to-one CLIP-style alignment by modeling a distribution over multiple positive image--text pairs, leading to more robust identity-level associations. The predicted image-to-text distribution for image $x_i$ is
\begin{equation}
p_{i,j}=\frac{\exp\!\big(s(f_\theta(x_i),\mathbf{u}_j)/\tau_{im}\big)}
{\sum_{k=1}^{B}\exp\!\big(s(f_\theta(x_i),\mathbf{u}_k)/\tau_{im}\big)},
\label{eq:iam_p}
\end{equation}
where $j$ indexes the candidate captions in the mini-batch and $\tau_{im}$ is a temperature controlling the sharpness of the distribution. The identity-conditioned soft target spreads probability mass over all captions of the anchor's identity,
\begin{equation}
q_{i,j}=
\begin{cases}
\alpha, & i=j\\[4pt]
\dfrac{1-\alpha}{N_{pos}}, & y_j=y_i,\; j\neq i\\[6pt]
0, & \text{otherwise}
\end{cases}
\label{eq:iam_q}
\end{equation}
where $N_{pos}$ is the number of captions in the mini-batch belonging to identity $y_i$ other than the anchor's own, and $\alpha\in(0,1)$ balances the weight assigned to the anchor pair against that shared among the remaining positives. Setting $\alpha=1$ recovers strict one-to-one matching. The loss is the KL divergence between prediction and target,
\begin{equation}
\mathcal{L}_{it}=\frac{1}{B}\sum_{i=1}^{B}\sum_{j=1}^{B}
p_{i,j}\log\frac{p_{i,j}}{q_{i,j}+\epsilon},
\label{eq:iam1}
\end{equation}
\begin{equation}
\mathcal{L}_{im}=\mathcal{L}_{it}+\mathcal{L}_{ti},
\label{eq:iam2}
\end{equation}
where $\epsilon$ is a small constant ensuring numerical stability and $\mathcal{L}_{ti}$ is defined symmetrically by swapping the roles of the two modalities in~\eqref{eq:iam_p}. Unlike CLIP's strict one-to-one matching, \eqref{eq:iam_q} distributes supervision across multiple positives of the same identity, which is precisely the identity-level behavior we later show to conflict with instance-level T2I supervision~(Sec.~\ref{sec:conflict}).

In addition, the Structure-preserving Loss regularizes the shared embedding space by maintaining the structural consistency of image representations:
\begin{equation}
\mathcal{L}_{sp}=\mathbb{E}\!\left[
-\log\frac{w_{ij^{*}}}
{w_{ij^{*}} + \sum_{k=1}^{N_{neg}}\exp\!\big(s(\mathbf{v}_i,\mathbf{v}_k)/\tau_{sp}\big)}
\right],
\label{eq:sp}
\end{equation}
where $w_{ij^{*}} = \exp\!\big(s(\mathbf{v}_i,\mathbf{v}_{j^{*}})/\tau_{sp}\big)$ denotes the exponential of the temperature-scaled similarity between anchor $i$ and instance $j^{*}$, $j^{*}=\arg\min_{j:\,y_j=y_i}s(\mathbf{v}_i,\mathbf{v}_j)$ is the hardest positive of the anchor, i.e. the same-identity image currently farthest from it. $N_{neg}=B-N_{pos}-1$ is the number of images in the mini-batch belonging to other identities, and $\tau_{sp}$ is a temperature. Whereas IAM operates across modalities, this term acts within the visual modality alone, preventing the image embeddings of one identity from drifting apart as they are pulled toward their captions. The overall image-text matching objective is
\begin{equation}
\mathcal{L}_{\text{match}}=\mathcal{L}_{im}+\mathcal{L}_{sp}.
\label{eq:im+sp}
\end{equation}

We show that incorporating textual supervision during vision encoder pre-training improves both I2I and subsequent T2I ReID performance, even though the vision encoder is frozen during the cross-modal alignment~(Sec.~\ref{sec:text_supervision}). Besides, ReText is capable of generalizing on multi-source training, which allows us to assess the impact of domain intersection between two stages~(Sec.~\ref{sec:domain})\footnote{The I2I and T2I datasets widely used for evaluating ReID intersect with each other.}.

\subsection{Stage 2: Cross-Modal Alignment}

In the second stage, the pre-trained vision encoder is frozen and integrated into the RDE pipeline. The training focuses exclusively on optimizing the text encoder to align textual representations with the fixed visual embedding space. This stage follows the standard RDE training protocol, ensuring effective cross-modal retrieval while preventing cross-task interference that could negatively impact image-based ReID performance. The text encoder is trained with the Triplet Alignment Loss (TAL), which relaxes the conventional Triplet Ranking Loss (TRL) by replacing the hard maximum with a log-sum-exp over all negatives. 
TAL is applied to each pair $(x_i,t_i)$ in a mini-batch of size $B$:
\begin{align}
\mathcal{L}_{tal}^{i}=& \Big[m-s^{+}_{i2t}(x_i)+\tau_{tal}\log\!\textstyle\sum_{j=1}^{B} q_{ij}\exp\!\big(s(\mathbf{v}_i,\mathbf{u}_j)/\tau_{tal}\big)\Big]_{+}\notag\\ +&\Big[m-s^{+}_{t2i}(t_i)+\tau_{tal}\log\!\textstyle\sum_{j=1}^{B} q_{ji}\exp\!\big(s(\mathbf{v}_j,\mathbf{u}_i)/\tau_{tal}\big)\Big]_{+},
\label{eq:tal}
\end{align}
where $[x]_{+}=\max(x,0)$ truncates negative values, $m$ is a positive margin, and $\tau_{tal}$ is a temperature controlling how sharply hard negatives are weighted. The binary correspondence label $l_{ij}\in\{0,1\}$ indicates whether caption $t_j$ matches image $x_i$, so its complement $q_{ij}=1-l_{ij}$ masks out positives and leaves only negatives inside the log-sum-exp. Since several images of one identity may co-occur in a mini-batch, the positive term is a weighted average over all of them,
\begin{equation}
s^{+}_{i2t}(x_i)=\sum_{j=1}^{B}\alpha_{ij}\,s(\mathbf{v}_i,\mathbf{u}_j),
\label{eq:tal_pos1}
\end{equation}
\begin{equation}
\alpha_{ij}=\frac{l_{ij}\exp\!\big(s(\mathbf{v}_i,\mathbf{u}_j)/\tau_{tal}\big)}
{\sum_{k=1}^{B}l_{ik}\exp\!\big(s(\mathbf{v}_i,\mathbf{u}_k)/\tau_{tal}\big)},
\label{eq:tal_pos2}
\end{equation}
with $s^{+}_{t2i}(t_i)$ defined symmetrically by swapping the roles of the two modalities. Since the weights $\alpha_{ij}$ are treated as constants during backpropagation, the gradient for each positive equals its weight, and with a small $\tau_{tal}$ it is dominated by the most similar positive, typically the paired image. Combined with random sampling, utilized in RDE, where a mini-batch rarely contains several images of one identity, this makes TAL a predominantly instance-level objective, making it effective for T2I retrieval. Using a larger $\tau_{tal}$ or identity-balanced sampling shifts TAL towards an identity-level objective. 

For convenience, we refer to the two-stage pipeline as UNI, with its first and second stages denoted as UNI (Stage 1) and UNI (Stage 2), respectively. Hereafter, RDE and ReText refer specifically to the corresponding methods, rather than to components of the pipeline studied in this work.

\section{Experimental Setup}
\subsection{Datasets and Metrics} 
\label{subsec: datasets}
We adopt CUHK03~\cite{cuhk03}, Market-1501~\cite{market1501}, and MSMT17~\cite{msmt17} for I2I ReID evaluation. The key difference among them is the number of camera views --- CUHK03 is collected from 2 camera views, Market-1501 --- from 6 cameras, and MSMT17 is the most challenging dataset, gathered from 15 cameras.

For T2I ReID evaluation, we use CUHK-PEDES~\cite{cuhk_pedes}, ICFG-PEDES~\cite{icfg_pedes}, and RSTPReid~\cite{rstpreid}, which provide natural language descriptions aligned with person images. Note that CUHK-PEDES is constructed using images from multiple I2I ReID datasets, including CUHK03 and Market-1501. Additionally, both ICFG-PEDES and RSTPReid are based on MSMT17 images. Such data overlaps across tasks are explicitly considered in our training and evaluation protocols to prevent data leakage and ensure reliable experimental comparisons. This applies to all experiments except those in which we intentionally mix in overlapping data to assess its impact.

We train ReText on I2I ReID datasets that do not overlap with the described benchmarks — ENTIRe-ID~\cite{enti}, IUSTPersonReID~\cite{iust}, CUHK-SYSU~\cite{cuhk_sysu}, PANDA~\cite{panda}, and OWD~\cite{owd}. For clarity, the mix of these 5 datasets will be referred to as the 5D-combo. Since a T2I ReID dataset is required to train ReText, we utilize SYNTH-PEDES~\cite{synth_pedes}, which contains 1 million images with text. Note that the SYNTH-PEDES captions are automatically generated pseudo-captions, and their quality was analyzed in ~\cite{synth_pedes}. This dataset is also employed in Stage 2 to pre-train the text encoder and projection before fine-tuning on the target benchmarks. During this pre-training, model selection is based on the validation split of ORBench~\cite{orbench}, which does not overlap with any evaluation benchmark.

We adopt the standard evaluation protocol for person ReID and report mean Average Precision~(mAP) and Rank-1 accuracy~(R1) for all I2I and T2I ReID experiments. These metrics are computed at the identity-level consistent with prior work.

\subsection{Implementation Details} 
We follow the original ReText~\cite{ReText} and RDE~\cite{RDE} training protocols, changing only the components required by each experiment.

\vspace{2pt}\noindent\textbf{Stage 1.} The vision encoder is a ViT-Base, initialized from self-supervised pre-training on LUPerson, and is accompanied by a momentum encoder of identical architecture updated by exponential moving averaging --- only the momentum encoder is used at inference. The text encoder is BERT-Base. We train UNI (Stage 1) for 100 epochs with AdamW, a learning rate of $10^{-5}$, a weight decay of $0.02$, and a warm-up over the first 10 epochs. Each mini-batch contains 96 images: 32 from multi-camera data, sampled with the PK strategy using $P_m=8$ identities and $K_m=4$ images per identity from different cameras, and 64 from single-camera data with $P_s=32$ and $K_s=2$, where camera diversity is not enforced. All images are resized to $256\times128$ and augmented with random cropping, horizontal flipping, Gaussian blurring, and random grayscale conversion. In the Identity-aware Matching Loss we set $\alpha=0.6$, in the Structure-preserving Loss $\tau_{sp}=0.1$. In the general setting, UNI (Stage 1) is trained on the 5D-combo as I2I ReID data and on SYNTH-PEDES as T2I ReID data. To assess the impact of different dataset combinations and matching objectives, it is retrained with the corresponding elements replaced. Training is performed on two NVIDIA GeForce RTX 3090 GPUs.

\vspace{2pt}\noindent\textbf{Stage 2.} The pre-trained vision encoder is frozen and integrated into UNI (Stage 2) in place of its CLIP visual branch, while the text branch follows the original design. A projection layer maps the two modalities into a shared embedding space. Gradients therefore update only the text encoder and the projection layer. I2I retrieval uses the vision encoder embeddings before projection, so Stage 2 training does not affect I2I results, while the projected embeddings are used only for T2I retrieval. Because the visual encoder is frozen, images are kept at the $256\times128$ resolution used during its pre-training rather than the $384\times128$ resolution of the original RDE. We train the text encoder for 60 epochs with Adam, a cosine learning-rate decay, an initial learning rate of $10^{-5}$ for the pre-trained parameters and $10^{-3}$ for the newly initialized ones, and a batch size of 64. Captions are tokenized to a maximum length of 168 and augmented with random masking, replacement, and removal of word tokens. Images undergo random horizontal flipping, random cropping with padding, and random erasing. Although the visual encoder is not updated, these image augmentations still diversify the visual embeddings the text encoder is aligned to, and are therefore retained. The Triplet Alignment Loss uses a margin $m=0.1$ and a temperature $\tau_{tal}=0.015$. Both similarity branches are kept: in the token selection embedding module, the selection ratio is $0.3$ for text tokens and $0.7$ for image patches, and the final similarity is the average of the global and token-selection scores. Training is performed on four NVIDIA GeForce RTX 2080 Ti GPUs.

\begin{table}[t]
\centering
\resizebox{\linewidth}{!}{
\begin{tabular}{l|cccc|c|cc}
\multirow{3}{*}{\textbf{Methods}} &
\multicolumn{4}{c|}{\textbf{I2I ReID}} &
\multirow{3}{*}{\makecell{\textbf{avg I2I} \\ \textbf{mAP}}} &
\multicolumn{2}{c}{\textbf{T2I ReID}} \\
 &
\multicolumn{2}{c}{\textbf{Market-1501}} &
\multicolumn{2}{c|}{\textbf{MSMT17}} &
 &
\multicolumn{2}{c}{\textbf{CUHK-PEDES}} \\
 &
R1 & mAP &
R1 & mAP &
 & R1 & mAP \\ \hline
RDE & 75.2 & 62.6 & 70.0 & 39.3 & 51.0 & 68.6 & 63.5 \\
UNI & 97.2 & 93.4 & 91.0 & 78.1 & 85.8 & 70.2 & 65.0 \\
\end{tabular}}
\caption{The comparison of T2I method RDE and the I2I method UNI.}
\label{tab:pretrain_impact_a}
\end{table}

\begin{table*}[t]
\centering
\resizebox{\textwidth}{!}{
\begin{tabular}{cc|cccccc|c|cccccc|c|c}
\multirow{3}{*}{\textbf{Stage 1}} & \multirow{3}{*}{\textbf{Stage 2}} & \multicolumn{6}{c|}{\textbf{I2I ReID}} & \multirow{3}{*}{\makecell{\textbf{avg I2I} \\ \textbf{mAP}}} & \multicolumn{6}{c|}{\textbf{T2I ReID}} & \multirow{3}{*}{\makecell{\textbf{avg T2I} \\ \textbf{mAP}}} & \multirow{3}{*}{\makecell{\textbf{avg} \\ \textbf{mAP}}} \\ 
 &  & \multicolumn{2}{c}{\textbf{CUHK03}} & \multicolumn{2}{c}{\textbf{Market-1501}} & \multicolumn{2}{c|}{\textbf{MSMT17}} &  & \multicolumn{2}{c}{\textbf{CUHK-PEDES}} & \multicolumn{2}{c}{\textbf{ICFG-PEDES}} & \multicolumn{2}{c|}{\textbf{RSTPReid}} &  &  \\ 
 &  & \multicolumn{1}{c}{R1} & \multicolumn{1}{c}{mAP} & \multicolumn{1}{c}{R1} & \multicolumn{1}{c}{mAP} & \multicolumn{1}{c}{R1} & mAP &  & \multicolumn{1}{c}{R1} & \multicolumn{1}{c}{mAP} & \multicolumn{1}{c}{R1} & \multicolumn{1}{c}{mAP} & \multicolumn{1}{c}{R1} & mAP &  &  \\ \hline
\multirow{2}{*}{\textbf{IAM}} & \textbf{IAM}
& \multirow{2}{*}{67.2}
& \multirow{2}{*}{68.5}
& \multirow{2}{*}{94.1}
& \multirow{2}{*}{86.1}
& \multirow{2}{*}{83.8}
& \multirow{2}{*}{65.1}
& \multirow{2}{*}{73.2}
& 72.6 & 66.7 & 63.1 & 39.1 & 61.2 & 49.6 & 51.8 & 62.5 \\
 & \textbf{TAL}
&  &  &  &  &  &  &
& 73.5 & 67.6 & 65.1 & 40.7 & 62.3 & 48.7 & 52.3 & 62.8 \\ \hline
\multirow{2}{*}{\textbf{TAL}} & \textbf{IAM}
& \multirow{2}{*}{64.3}
& \multirow{2}{*}{64.7}
& \multirow{2}{*}{94.0}
& \multirow{2}{*}{85.2}
& \multirow{2}{*}{81.1}
& \multirow{2}{*}{60.0}
& \multirow{2}{*}{70.0}
& \multicolumn{1}{c}{72.6} & 66.4
& \multicolumn{1}{c}{64.0} & 39.2
& \multicolumn{1}{c}{61.7} & 48.8 & 51.5 & 60.8 \\
 & \textbf{TAL}
&  &  &  &  &  &  &
& 73.8 & 67.7
& 65.9 & 40.7
& 62.5 & 49.8 & 52.7 & 61.4 \\
\end{tabular}}
\caption{Effect of specially designed loss functions for effective processing of different query modalities.}
\label{tab:loss_impact}
\end{table*}

\section{Results and Analysis}
\label{sec:results}

\subsection{Can the T2I method directly solve the I2I problem?}
\label{sec:rde_i2i}
\vspace{2pt}\noindent\textbf{Setup.} We investigate whether a T2I ReID method can directly address the I2I ReID task. To this end, we use RDE as the T2I baseline and compare its I2I performance with that of UNI (Stage 1). 
There is an opportunity to train both T2I and I2I methods using the same image data, thanks to the annotated multi-camera datasets provided in \cite{ReText}. As a result, RDE and UNI (Stage 1) are trained on the training splits of the text-supervised MSMT17 and Market-1501 datasets and evaluated on the corresponding test splits. Both methods use the same ViT-Base backbone, visual initialization, identity sampler, and a training budget of 60 epochs, so they differ only in the training objective. In both cases, we use the vision encoder to extract image embeddings, which are then evaluated on image-based ReID benchmarks.

\vspace{2pt}\noindent\textbf{Results.} As shown in the I2I part of Table~\ref{tab:pretrain_impact_a}, the RDE substantially degrades in performance compared to the method explicitly optimized for image-based ReID. These findings indicate that cross-modal supervision alone is insufficient to produce identity-discriminative representations required for competitive I2I performance. 

\begin{figure*}[t]
  \centering
  \includegraphics[width=0.95\textwidth]{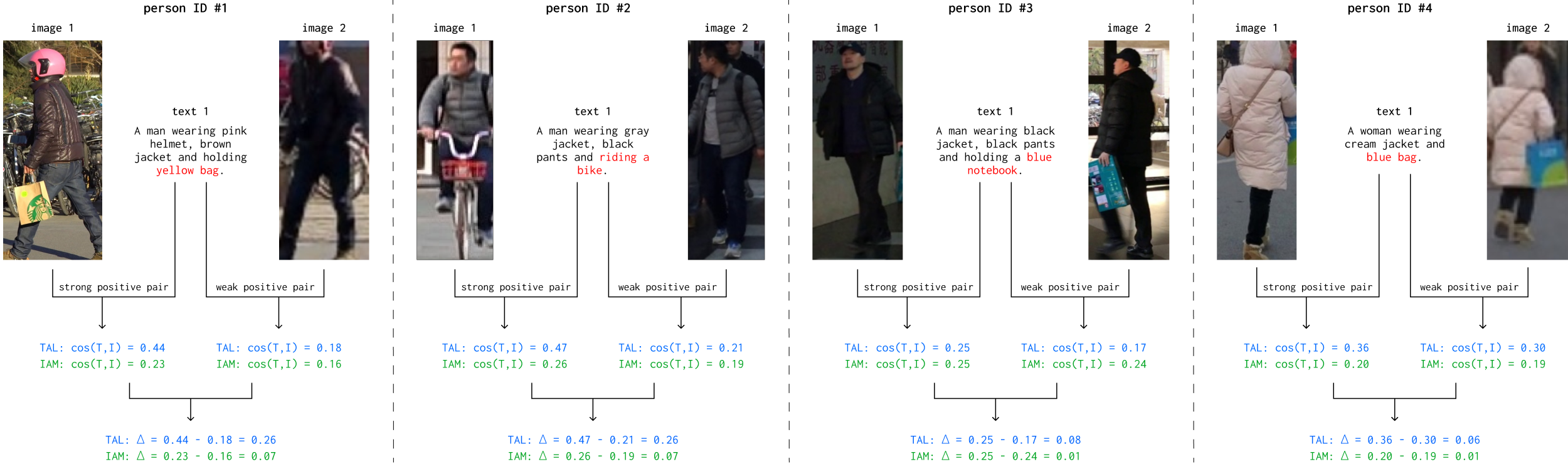}
  \caption{Instance-level sensitivity of IAM- and TAL-trained spaces. For each caption describing one specific image, we measure its cosine similarity to the described instance and to another image of the same identity in which the mentioned attribute is absent, and report the gap $\Updelta$. In the TAL-trained space, $\Updelta$ is consistently large: the caption is markedly closer to the instance it actually describes. In the IAM-trained space, $\Updelta$ collapses toward zero, meaning the two instances of one identity are nearly indistinguishable to the text. This confirms that IAM enforces identity-level invariance, whereas TAL preserves the instance-level distinctions required for T2I retrieval.}
  \label{fig:losses}
\end{figure*}

\subsection{Does I2I pre-training provide a better visual anchor for T2I ReID than T2I pre-training?}
\label{sec:anchor}
To identify the most effective pre-training method for fine-tuning on the T2I ReID task, we compare vision encoders trained with I2I and T2I objectives as anchors for cross-modal alignment.

\vspace{2pt}\noindent\textbf{Setup.} We reuse the vision encoders of RDE and UNI~(Stage 1) trained on MSMT17 in Section~\ref{sec:rde_i2i}. Each encoder is frozen, and the text encoder with projection is trained on CUHK-PEDES following the UNI (Stage 2) protocol, which is similar to RDE's. The MSMT17 checkpoints are chosen because CUHK-PEDES does not overlap with MSMT17, ensuring that the T2I evaluation is free of data leakage.

\begin{table*}[t]
\centering
\resizebox{\textwidth}{!}{
\begin{tabular}{l|cccccc|c|cccccc|c|c}
\multirow{3}{*}{\textbf{Methods}} &
\multicolumn{6}{c|}{\textbf{I2I ReID}} &
\multirow{3}{*}{\makecell{\textbf{avg I2I} \\ \textbf{mAP}}} &
\multicolumn{6}{c|}{\textbf{T2I ReID}} &
\multirow{3}{*}{\makecell{\textbf{avg T2I} \\ \textbf{mAP}}} &
\multirow{3}{*}{\makecell{\textbf{avg} \\ \textbf{mAP}}} \\
 &
\multicolumn{2}{c}{\textbf{CUHK03}} &
\multicolumn{2}{c}{\textbf{Market-1501}} &
\multicolumn{2}{c|}{\textbf{MSMT17}} &
 &
\multicolumn{2}{c}{\textbf{CUHK-PEDES}} &
\multicolumn{2}{c}{\textbf{ICFG-PEDES}} &
\multicolumn{2}{c|}{\textbf{RSTPReid}} &
 &
\\
 &
R1 & mAP &
R1 & mAP &
R1 & mAP &
 &
R1 & mAP &
R1 & mAP &
R1 & mAP &
 &
\\ \hline
UNI$_{\text{trained}}$ & 42.6 & 42.6 & 81.2 & 70.3 & 71.5 & 41.4 & 51.4 & 70.8 & 64.7 & 63.3 & 38.5 & 60.3 & 45.2 & 49.5 & 50.5\\
UNI$_{\text{last-}4}$ & 52.8 & 53.6 & 86.7 & 78.3 & 76.9 & 50.2 & 60.7 & 68.9 & 63.6 & 62.3 & 37.9 & 60.1 & 45.2 & 48.9 & 54.8\\
UNI$_{\text{last-}2}$ & 59.2 & 60.7 & 87.7 & 81.7 & 79.9 & 56.0 & 66.1 & 68.2 & 63.5 & 60.9 & 37.8 & 58.7 & 45.5 & 48.9 & 57.5\\
UNI$_{\text{lr}/10}$ & 60.1 & 61.2 & 86.0 & 79.8 & 79.1 & 55.3 & 65.4 & 69.2 & 65.2 & 62.1 & 39.8 & 57.7 & 46.4 & 50.5 & 58.0\\
UNI$_{\text{lr}/100}$ & 67.6 & 68.4 & 86.1 & 81.9 & 83.0 & 64.1 & 71.5 & 67.2 & 64.1 & 60.7 & 40.1 & 55.4 & 46.4 & 50.2 & 60.9\\
UNI & 67.2 & 68.5 & 94.1 & 86.1 & 83.8 & 65.1 & 73.2 & 68.1 & 62.9 & 61.3 & 37.7 & 60.1 & 45.7 & 48.8 & 61.0\\
\end{tabular}}
\caption{I2I and T2I performance of UNI under different strategies for training the vision encoder in Stage 2, ordered from full adaptation to full freezing: trainable encoder (UNI$_{\text{trained}}$), learning rate reduced by 10$\times$ and 100$\times$ (UNI$_{\text{lr}/10}$, UNI$_{\text{lr}/100}$), updating only the last 4 or 2 blocks (UNI$_{\text{last-4}}$, UNI$_{\text{last-2}}$), and frozen encoder (UNI).}
\label{tab:freezing_impact}
\end{table*}

\vspace{2pt}\noindent\textbf{Results.} As shown in the T2I part of Table~\ref{tab:pretrain_impact_a}, the encoder pre-trained with the I2I objective provides a better anchor for cross-modal alignment than the one pre-trained with the T2I objective. Thus, under identical setups, the identity-level structure learned for I2I ReID is not only preserved by freezing, but also transfers to T2I retrieval better than the representation learned with the T2I objective itself. We attribute this to the cross-camera invariance enforced by the I2I objective, which yields visual embeddings that are robust to viewpoint and appearance changes and are therefore easier to align with textual descriptions of the same person.

\subsection{Do image- and text-based ReID objectives conflict during joint optimization?}
\label{sec:conflict}
We check this hypothesis within the studied two-stage pipeline. We vary the text-image matching loss applied to each stage to assess whether identity-level I2I and instance-level T2I objectives are interchangeable. The Identity-aware Matching~(IAM) loss, proposed by ReText authors, unlike one-to-one matching in CLIP, learns a distribution over multiple positive image-text pairs, enabling more flexible and identity-aware alignment. In contrast, the Triplet Alignment Loss~(TAL), developed by RDE authors, operates on individual image-text pairs and, unlike standard Triplet Ranking Loss (TRL), relaxes the optimization of the hardest negatives to all negatives with an upper bound.

\vspace{2pt}\noindent\textbf{Setup.} In UNI (Stage 1), we pre-train one vision encoder with the IAM loss and another with the TAL loss. Each encoder is then integrated into UNI (Stage 2), and the text encoder is tuned using both loss functions separately.

\vspace{2pt}\noindent\textbf{Results.} The IAM loss leads to stronger performance in image-based ReID, whereas the TAL is more effective for T2I ReID~(Tab.~\ref{tab:loss_impact}). The swap of these objectives between stages results in a degradation of the quality. These results indicate that, despite their conceptual similarity, I2I and T2I ReID require different optimization strategies. IAM promotes identity-level consistency by aggregating multiple positives per identity, whereas TAL enforces instance-level ranking via triplet comparisons, better capturing fine-grained text-image alignment. This distinction is directly visible in the sensitivity of each space to instance-specific attributes. Fig.~\ref{fig:losses} shows that the TAL-trained space keeps a caption grounded in the one image it describes, separating it from another image of the same identity used as a distractor. The IAM-trained space instead makes the two instances nearly indistinguishable to the text, despite visible attribute differences.

\begin{figure*}[t]
  \centering
  \includegraphics[width=0.92\textwidth]{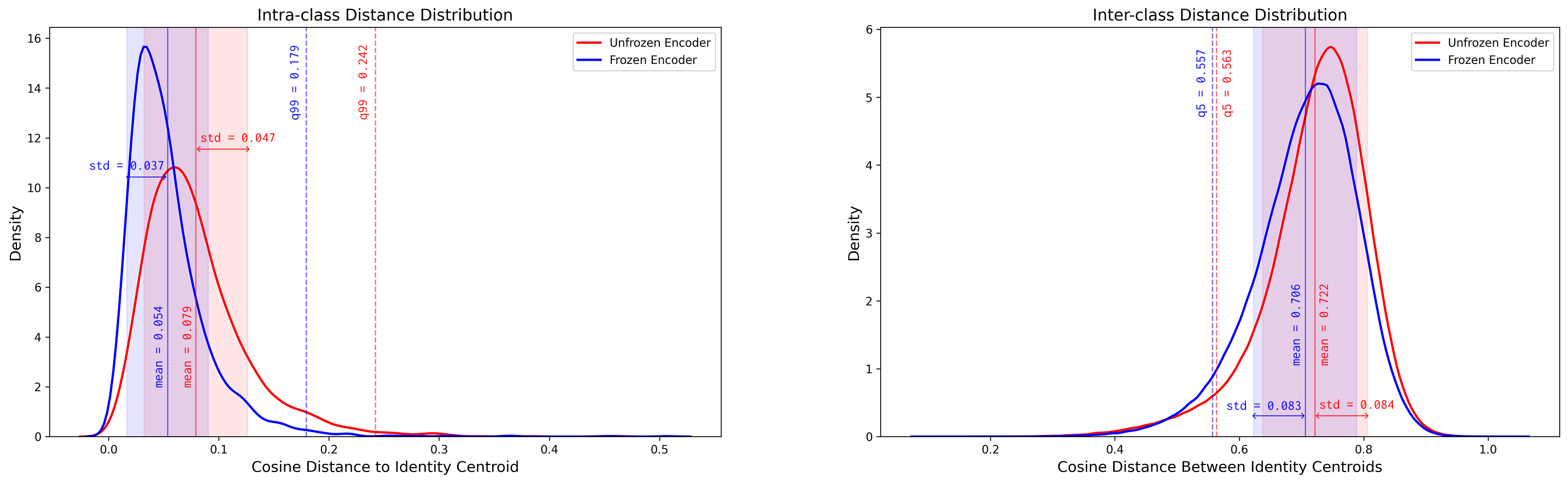}
  \caption{Effect of vision encoder freezing on embedding geometry. We compare intra-class (left) and inter-class (right) cosine-distance distributions for a frozen I2I-pretrained encoder and an encoder fine-tuned on T2I data with unfrozen weights. Fine-tuning on T2I inflates intra-class distances, indicating that identity clusters become less compact and cross-view invariance degrades. Inter-class separation between identity centroids is largely preserved across both settings.}
  \label{fig:geometry}
\end{figure*}

\subsection{Can a frozen vision encoder preserve I2I performance while serving as an effective anchor for T2I ReID?}
\label{sec:freeze}
Since there is a conflict between the two ReID tasks and their objectives, we assume that stabilizing the vision encoder after its I2I pre-training can help us align the text embedding space to it and obtain a strong unified model.

\vspace{2pt}\noindent\textbf{Setup.} We compare several versions of UNI that differ only in how the vision encoder is trained in Stage 2: fully frozen, fully trainable, trained with a learning rate reduced by 10$\times$ or 100$\times$, and partially frozen, with only the last 2 or 4 blocks updated. In all cases, UNI (Stage 1) is pre-trained on 5D-combo and SYNTH-PEDES, and Stage 2 is trained on SYNTH-PEDES without domain fine-tuning. Thus, each variant is a single model evaluated on all six benchmarks in the generalization setting.

\begin{figure}[t]
  \centering
  \includegraphics[width=\linewidth]{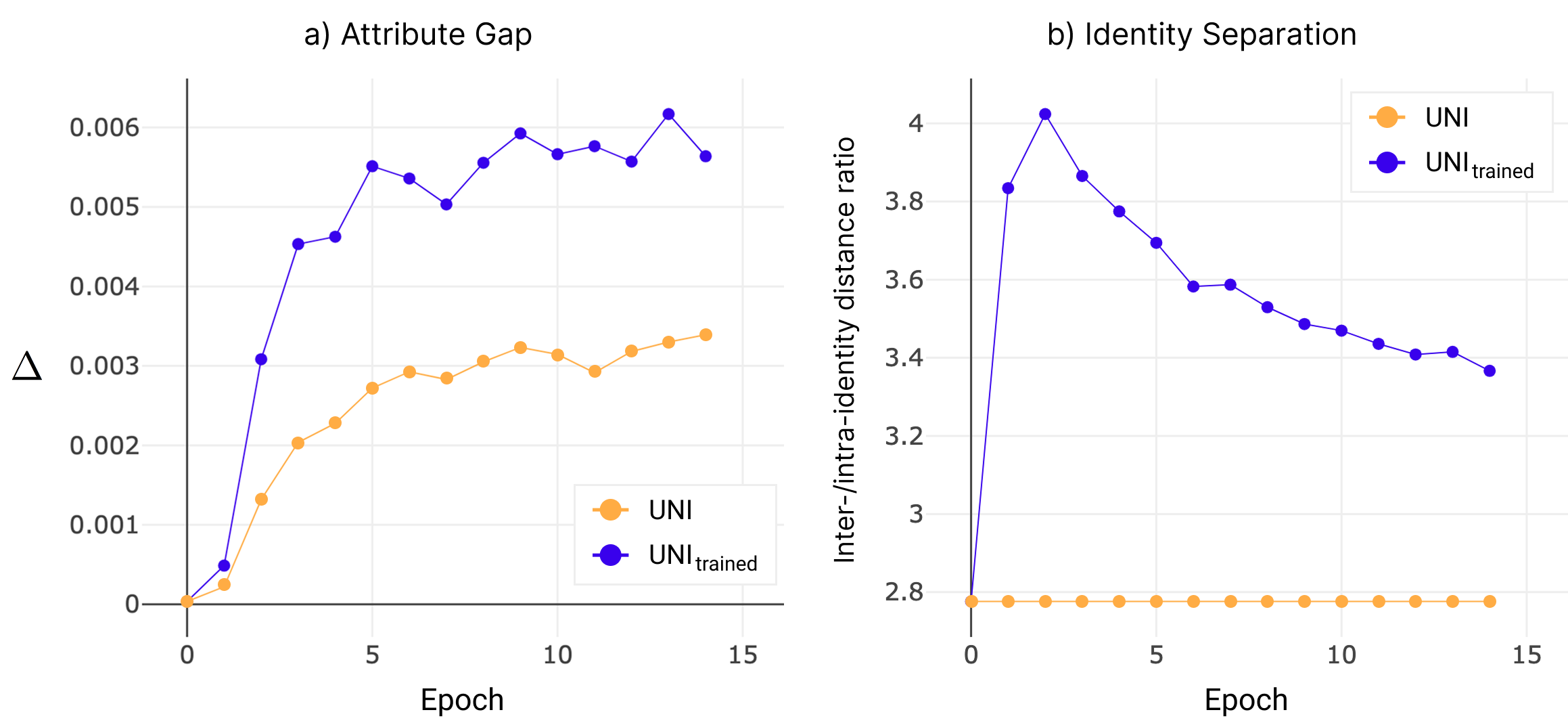}
  \caption{Attribute gap and identity separation during Stage 2 training with frozen (UNI) and trainable (UNI$_{\text{trained}}$) vision encoders. (a) The attribute gap $\Updelta$, computed in the shared image--text space, averaged over captions. (b) The ratio of mean inter-identity to mean intra-identity cosine distances. All metrics are measured on the ORBench.}
  \label{fig:gap}
\end{figure}

\vspace{2pt}\noindent\textbf{Results.} As shown in Table~\ref{tab:freezing_impact}, the more the encoder adapts, the larger the I2I degradation, while the T2I gain does not exceed 1.7 average mAP. The overall average mAP increases monotonically as the encoder adapts less, and the 100$\times$ reduced learning rate, under which the encoder barely changes, nearly matches full freezing (60.9 vs. 61.0 average mAP). Thus, full freezing is the limiting case of these stabilization strategies. We keep it as the default, since it preserves I2I performance by construction and introduces no additional hyperparameter.

Fig.~\ref{fig:geometry} visualizes the embedding distributions obtained by UNI with and without freezing the vision encoder on its second stage. As a result, the fine-tuning on T2I data shifts the intra-class distance distribution to the right, meaning images of the same identity, previously collapsed into a tight cluster by the frozen encoder, spread apart once the encoder is updated to satisfy the T2I objective. The inter-class distance distribution, in contrast, stays nearly the same in both settings. So the T2I objective separates the compactness within a single identity rather than confusing different identities.

To trace how this effect develops, Fig.~\ref{fig:gap} shows the attribute gap $\Updelta$ and identity separation throughout Stage 2 training. The attribute gap~\eqref{eq:delta} is the difference between the similarity of a caption to the image it describes and its similarity to other images of the same identity, averaged over captions, so its growth indicates stronger instance-level grounding. Identity separation is the ratio of the mean inter-identity to the mean intra-identity cosine distance between the vision encoder embeddings before projection, so higher values correspond to more compact and better separated identity clusters. With the trainable encoder, identity separation first increases as the encoder adapts to the training domain, but then decreases steadily. At the same time, $\Updelta$ keeps growing, so the T2I objective gradually loosens identity clusters as it strengthens instance-level grounding. With the frozen encoder, $\Updelta$ grows more slowly, since only the text side can adapt, and identity separation remains unchanged by construction.

\begin{table*}[t]
\centering
\resizebox{\textwidth}{!}{
\begin{tabular}{l|cccccc|c|cccccc|c|c}
\multirow{3}{*}{\textbf{Methods}} &
\multicolumn{6}{c|}{\textbf{I2I ReID}} &
\multirow{3}{*}{\makecell{\textbf{avg I2I} \\ \textbf{mAP}}} &
\multicolumn{6}{c|}{\textbf{T2I ReID}} &
\multirow{3}{*}{\makecell{\textbf{avg T2I} \\ \textbf{mAP}}} &
\multirow{3}{*}{\makecell{\textbf{avg} \\ \textbf{mAP}}} \\
 &
\multicolumn{2}{c}{\textbf{CUHK03}} &
\multicolumn{2}{c}{\textbf{Market-1501}} &
\multicolumn{2}{c|}{\textbf{MSMT17}} &
 &
\multicolumn{2}{c}{\textbf{CUHK-PEDES}} &
\multicolumn{2}{c}{\textbf{ICFG-PEDES}} &
\multicolumn{2}{c|}{\textbf{RSTPReid}} &
 &
\\
 &
R1 & mAP &
R1 & mAP &
R1 & mAP &
 &
R1 & mAP &
R1 & mAP &
R1 & mAP &
 &
\\ \hline
UNI$_{\text{LUPerson-NL}}$ & 61.6 & 62.7 & 92.9 & 83.2 & 82.2 & 62.1 & 69.3 & 72.2 & 66.3 & 62.7 & 37.7 & 60.8 & 47.2 & 50.4 & 59.9 \\
UNI$_{\text{SYNTH-PEDES wo text}}$ & 63.3 & 62.7 & 90.0 & 77.1 & 79.2 & 56.9 & 65.6 & 72.9 & 67.0 & 64.8 & 40.2 & 61.8 & 47.4 & 51.5 & 58.6\\  
UNI$_{\text{SYNTH-PEDES}}$ & 67.2 & 68.5 & 94.1 & 86.1 & 83.8 & 65.1 & 73.2 & 73.5 & 67.6 & 65.1 & 40.7 & 62.3 & 48.7 & 52.3 & 62.8 \\
\end{tabular}}
\caption{Effect of textual supervision during vision encoder pre-training. We compare three variants of UNI that differ only in the additional data used at the first stage. UNI (LUPerson-NL) uses text-free LUPerson-NL, while UNI~(SYNTH-PEDES wo text) and UNI (SYNTH-PEDES) use text-free and text-based SYNTH-PEDES, respectively.}
\label{tab:pretrain_impact_b}
\end{table*}

\begin{table}[t]
\centering
\resizebox{\linewidth}{!}{
\begin{tabular}{l|cccccc|c}
\multirow{3}{*}{\textbf{Methods}} &
\multicolumn{6}{c|}{\textbf{T2I ReID}} &
\multirow{3}{*}{\makecell{\textbf{avg T2I} \\ \textbf{mAP}}} \\
 &
\multicolumn{2}{c}{\textbf{CUHK-PEDES}} &
\multicolumn{2}{c}{\textbf{ICFG-PEDES}} &
\multicolumn{2}{c|}{\textbf{RSTPReid}} &
\\
 &
R1 & mAP &
R1 & mAP &
R1 & mAP &
\\ 
\hline
\multicolumn{8}{c}{\textbf{Generalization}} \\ \hline
RDE & 70.3 & 63.5 & 61.7 & 36.7 & 59.4 & 45.3 & 48.5 \\
UNI$_{\text{trained}}$ & 70.8 & 64.7 & 63.3 & 38.5 & 60.3 & 45.2 & 49.5 \\
UNI & 68.1 & 62.9 & 61.3 & 37.7 & 60.1 & 45.7 & 48.8 \\
\hline
\multicolumn{8}{c}{\textbf{Domain Fine-Tuning}} \\ \hline
RDE & 77.9 & 70.0 & 69.8 & 42.7 & 70.2 & 56.1 & 56.3 \\
UNI$_{\text{trained}}$ & 78.6 & 71.4 & 71.4 & 45.5 & 71.9 & 56.9 & 57.9 \\
UNI & 75.2 & 68.6 & 67.6 & 42.6 & 68.0 & 53.9 & 55.0 \\
\end{tabular}}
\caption{Transition from the T2I method RDE to UNI. UNI$_{\text{trained}}$ adds I2I pre-training of the vision encoder~(Stage 1), which remains trainable in Stage 2, while UNI additionally freezes it in Stage 2. All models are pre-trained on SYNTH-PEDES and evaluated without (generalization) and with domain fine-tuning.}
\label{tab:pretrain_impact_c}
\end{table}

\begin{table*}[t]
\centering
\resizebox{\textwidth}{!}{
\begin{tabular}{l|cc|ccc|cc|ccc}
\multirow{2}{*}{\textbf{Configuration}} &
\multicolumn{2}{c|}{\textbf{Stage 1 Data}} &
\multicolumn{3}{c|}{\textbf{I2I ReID (mAP)}} &
\multicolumn{2}{c|}{\textbf{Stage 2 Data}} &
\multicolumn{3}{c}{\textbf{T2I ReID (mAP)}} \\
 & I2I branch & T2I branch & CUHK03 & Market-1501 & MSMT17 & Pre-training & Fine-tuning & CUHK-P. & ICFG-P. & RSTPReid \\ \hline
\multirow{3}{*}{UNI} & \multirow{3}{*}{5D-combo} & \multirow{3}{*}{SYNTH-P.} & \multirow{3}{*}{68.5} & \multirow{3}{*}{86.1} & \multirow{3}{*}{65.1} & \multirow{3}{*}{SYNTH-P.} & CUHK-P. & 68.6 & — & — \\
 & & & & & & & ICFG-P. & — & 42.6 & — \\
 & & & & & & & RSTPReid & — & — & 53.9 \\ \hline
\multicolumn{11}{c}{\textbf{Text-based domain incorporation}} \\ \hline
\multirow{2}{*}{UNI$_{\text{CUHK-PEDES}}$} & \multirow{6}{*}{5D-combo} & CUHK-P. & 84.9 & 91.6 & 64.9 & \multirow{6}{*}{SYNTH-P.} & \multirow{2}{*}{CUHK-P.} & 68.0 & — & — \\
 & & SYNTH-P. + CUHK-P. & 85.7 & 92.5 & 66.7 & & & 68.5 & — & — \\ \cline{1-1} \cline{3-6} \cline{8-11}
\multirow{2}{*}{UNI$_{\text{ICFG-PEDES}}$} & & ICFG-P. & 65.3 & 84.1 & 75.2 & & \multirow{2}{*}{ICFG-P.} & — & 42.7 & — \\
 & & SYNTH-P. + ICFG-P. & 65.8 & 85.1 & 76.3 & & & — & 42.7 & — \\ \cline{1-1} \cline{3-6} \cline{8-11}
\multirow{2}{*}{UNI$_{\text{RSTPReid}}$} & & RSTPReid & 66.8 & 84.9 & 78.0 & & \multirow{2}{*}{RSTPReid} & — & — & 49.7 \\
 & & SYNTH-P. + RSTPReid & 66.9 & 85.7 & 80.0 & & & — & — & 50.0 \\ \hline
\multicolumn{11}{c}{\textbf{Image-based domain incorporation}} \\ \hline
\multirow{3}{*}{UNI$_{\text{C3, M, MS}}$} & \multirow{3}{*}{\makecell{CUHK03 \\ Market-1501 \\ MSMT17}} & \multirow{3}{*}{SYNTH-P.} & \multirow{3}{*}{88.7} & \multirow{3}{*}{92.7} & \multirow{3}{*}{76.9} & \multirow{3}{*}{SYNTH-P.} & CUHK-P. & 67.4 & — & — \\
 & & & & & & & ICFG-P. & — & 42.1 & — \\
 & & & & & & & RSTPReid & — & — & 51.6 \\
\end{tabular}}
\caption[Domain incorporation configurations]{Training configurations and mAP results of the domain incorporation experiments. In UNI$_{\text{CUHK-PEDES}}$, UNI$_{\text{ICFG-PEDES}}$, and UNI$_{\text{RSTPReid}}$, the subscript refers to the T2I dataset used in Stage 1, either alone or with SYNTH-PEDES. In UNI$_{\text{C3, M, MS}}$, the I2I branch uses CUHK03 (C3), Market-1501 (M), and MSMT17 (MS) instead of the 5D-combo dataset. Merged I2I cells denote a single vision encoder shared by the corresponding rows. In Stage 2, we train a separate text encoder with a projection on top of the frozen vision encoder for each T2I benchmark, while I2I retrieval uses the vision encoder embeddings before projection.}
\label{tab:domain_configs}
\end{table*}

\subsection{How does textual supervision during vision encoder pre-training affect T2I ReID performance?} 
\label{sec:text_supervision}
\vspace{2pt}\noindent\textbf{Setup.} To assess the impact of textual supervision, we replace the T2I SYNTH-PEDES dataset in UNI (Stage 1) with the I2I LUPerson-NL~\cite{luperson} dataset. Since the LUPerson-NL dataset is not annotated with text, it is utilized under the multi-camera I2I ReID branch. Since SYNTH-PEDES is a text-annotated subset of LUPerson-NL, we also train UNI~(Stage 1) on the SYNTH-PEDES images without captions. This model shares the images with the text-supervised one and differs from it only in the absence of textual supervision, which allows us to separate the effect of textual supervision from that of the image data. We train three models on the 5D-combo with the additional LUPerson-NL, SYNTH-PEDES without captions, and full SYNTH-PEDES datasets. Then, each vision encoder is frozen, and the text encoder with projection in UNI (Stage 2) is fine-tuned on the domain sets to assess T2I ReID capabilities.

\vspace{2pt}\noindent\textbf{Results.} Table~\ref{tab:pretrain_impact_b} shows the impact of textual anchoring during vision encoder pre-training on downstream I2I and T2I ReID performance.

In T2I retrieval, the model trained with captions outperforms the one trained on the same images without them across three benchmarks. Notably, the model pre-trained on SYNTH-PEDES even without captions outperforms the one pre-trained on the full LUPerson-NL in T2I retrieval. We attribute this to the filtration process applied to LUPerson-NL when constructing SYNTH-PEDES, which likely results in cleaner identity labels.

In I2I retrieval, the number of pre-training images is essential when captions are absent --- the model trained on LUPerson-NL performs similarly or better than one trained on the smaller SYNTH-PEDES without captions. However, once captions are added, SYNTH-PEDES becomes the best pre-training option for I2I retrieval as well, despite containing fewer images than LUPerson-NL. This is consistent with the observation of ReText~\cite{ReText} that textual descriptions, used as auxiliary supervision, enrich visual representations with semantic cues that are not available from images alone.

\subsection{How do I2I pre-training and encoder freezing affect generalization in the T2I task?}
\label{sec:pretrain}
\vspace{2pt}\noindent\textbf{Setup.} We trace the transition from a T2I-only method to UNI in two steps. First, RDE is replaced by UNI$_{\text{trained}}$, which adds I2I pre-training of the vision encoder in Stage 1 while keeping it trainable in Stage 2. Then, the encoder is frozen, resulting in UNI. All models use SYNTH-PEDES as T2I data and are evaluated under two protocols: 1) generalization, which excludes training data from target benchmarks, and 2) domain fine-tuning, which involves additional fine-tuning on the training split of each benchmark. Note that the first step changes not only the pre-training objective but also the training data and the visual initialization. A comparison in which only the pre-training objective differs is provided in Section~\ref{sec:anchor}.

\vspace{2pt}\noindent\textbf{Results.} As shown in Table~\ref{tab:pretrain_impact_c}, adding I2I pre-training improves T2I performance under both protocols. Freezing the encoder then costs only 0.7 points without target-domain data, but 2.9 points after domain fine-tuning, since the frozen encoder cannot adapt to the target domain. In other words, the more the model is adapted to the target domain, the larger the cost of freezing becomes, while preserving the I2I performance that degrades sharply when the encoder is trained~(Table~\ref{tab:freezing_impact}).

\subsection{Does incorporating domain information during vision encoder pre-training improve fine-tuning abilities on I2I and T2I datasets?}
\label{sec:domain}
We conduct two configurations to incorporate domain information into the vision encoder pre-training.

\begin{table*}[t]
\centering
\resizebox{\textwidth}{!}{
\begin{tabular}{l|cccccc|c|l|cccccc|c}
\multirow{3}{*}{\textbf{Methods}} & \multicolumn{6}{c|}{\textbf{I2I ReID}} & \multirow{3}{*}{\makecell{\textbf{avg I2I} \\ \textbf{mAP}}} & \multirow{3}{*}{\textbf{Methods}} & \multicolumn{6}{c}{\textbf{T2I ReID}} & \multirow{3}{*}{\makecell{\textbf{avg T2I} \\ \textbf{mAP}}}\\
 & \multicolumn{2}{c}{\textbf{CUHK03}} & \multicolumn{2}{c}{\textbf{Market-1501}} & \multicolumn{2}{c|}{\textbf{MSMT17}} & & & \multicolumn{2}{c}{\textbf{CUHK-PEDES}} & \multicolumn{2}{c}{\textbf{ICFG-PEDES}} & \multicolumn{2}{c|}{\textbf{RSTPReid}} &\\
 
 & R1 & mAP & R1 & mAP & R1 & mAP & & & R1 & mAP & R1 & mAP & R1 & mAP &\\ \hline
\multicolumn{8}{c|}{\textbf{I2I-only Methods}} & \multicolumn{8}{c}{\textbf{T2I-only Methods}} \\ \hline
\textbf{TransReID}~\cite{transreid} & — & — & 94.4 & 86.8 & 81.8 & 61.0 & 73.9* & \textbf{IRRA}~\cite{irra} & 73.4 & 66.1 & 63.5 & 38.1 & 60.2 & 47.2 & 50.5\\
\textbf{PASS}~\cite{pass} & — & — & 96.8 & 93.0 & 88.2 & 71.8 & 82.4* & \textbf{RDE}~\cite{RDE} & 75.9 & 67.6 & 67.7 & 40.1 & 65.4 & 50.9 & 52.9\\
\textbf{CLIP-ReID}~\cite{clipreid} & — & — & 95.5 & 89.6 & 88.7 & 73.4 & 81.5* & \textbf{RaSa}~\cite{rasa} & 76.5 & 69.4 & 65.3 & 41.3 & 67.0 & 52.3 & 54.3\\
\textbf{ReMix}~\cite{remix} & — & — & 96.2 & 89.8 & 84.8 & 63.9 & 76.9* & \textbf{MARS}~\cite{mars} & 77.6 & 71.7 & 67.6 & 44.9 & 67.6 & 52.9 & 56.5\\
\textbf{DynaMix}~\cite{dynamix} & — & — & 97.4 & 93.8 & 90.4 & 76.3 & 85.1* & \textbf{CFAM}~\cite{cfam} & 75.6 & 67.3 & 65.4 & 39.4 & 62.5 & 49.5 & 52.1\\ \hline
\multicolumn{16}{c}{\textbf{Unified Methods}} \\ \hline
\textbf{IRM}~\cite{irm} & 86.5 & 85.4 & 96.5 & 93.5 & 86.9 & 72.4 & 83.8 & \textbf{IRM}~\cite{irm} & 74.2 & 66.5 & — & — & — & — & —\\
\textbf{HPL}~\cite{hpl} & — & — & 96.0 & 89.8 & 91.0 & 79.0 & 84.4* & \textbf{HPL}~\cite{hpl} & 76.3 & 70.9 & 66.6 & 44.1 & 64.0 & 53.1 & 56.1\\
\textbf{ReText}~\cite{ReText} & — & — & 97.4 & 93.8 & 91.4 & 78.7 & 86.3* & \textbf{ReText}~\cite{ReText} & 34.7 & 37.6 & 30.2 & 22.9 & 24.8 & 24.9 & 28.5\\
\textbf{UNI} & 89.6 & 88.7 & 96.7 & 92.7 & 90.1 & 76.9 & 86.1 & \textbf{UNI} & 75.2 & 68.6 & 67.6 & 42.6 & 68.0 & 53.9 & 55.0
\end{tabular}}
\caption{Comparison of I2I-only, T2I-only, and unified ReID methods on standard image-based and text-based benchmarks. * means the average metric calculated using only two available datasets, making it non-comparable to average metrics based on three datasets. Although ReText was proposed for only the I2I task, since it contains a text encoder, its metrics on T2I datasets are also included in the table. In contrast to Tabs.~\ref{tab:pretrain_impact_a}-\ref{tab:domain_configs}, the results on I2I datasets are obtained using the UNI (Stage 1) checkpoint trained on the corresponding training splits of the same datasets.}
\label{tab:sota}
\end{table*}

\vspace{2pt}\noindent\textbf{Text-based domain incorporation.} \textit{Setup.} We replace SYNTH-PEDES in UNI's T2I ReID branch with CUHK-PEDES, ICFG-PEDES, and RSTPReid separately, and obtain three vision encoders. Then, we fine-tune the text encoder and a projection layer in UNI (Stage 2) on the same corresponding datasets. Also, we prepare another three vision encoders with the addition of SYNTH-PEDES to evaluate its importance in vision encoder pre-training. All configurations are summarized in Table~\ref{tab:domain_configs}.

\textit{Results.} Table~\ref{tab:domain_configs} shows that incorporating CUHK-PEDES leads to a consistent performance improvement on CUHK03 and Market-1501 datasets that overlap with its visual samples. Similarly, pre-training with ICFG-PEDES or RSTPReid boosts the performance on MSMT17. This confirms that domain alignment between pre-training and downstream data positively impacts I2I retrieval. However, reducing the text-based data scale during pre-training negatively affects T2I metrics. Furthermore, even adding SYNTH-PEDES does not significantly improve the situation. This shows that data quantity outweighs domain overlap for the T2I task, and that it is more effective to focus pre-training on a single text-based dataset. 

\vspace{2pt}\noindent\textbf{Image-based domain incorporation.} \textit{Setup.} We replace the 5D-combo in the I2I branch with the combination of MSMT17, Market-1501, and CUHK03, whose images overlap with the downstream T2I benchmarks. We then freeze the resulting vision encoder and use it in UNI (Stage 2) to assess whether this domain overlap facilitates cross-modal alignment.

\textit{Results.} Table~\ref{tab:domain_configs} shows that adding domain-specific image-based information improves I2I retrieval but hurts text-based search. In summary, the quantity of image- and text-based data is more important for the T2I task, while domain overlap matters more for the I2I task. Furthermore, the drop in metrics after mixing domains confirms that ReText's IAM loss optimization is not sufficient for the T2I task.

\begin{figure*}[t]
  \centering
  \includegraphics[width=0.98\textwidth]{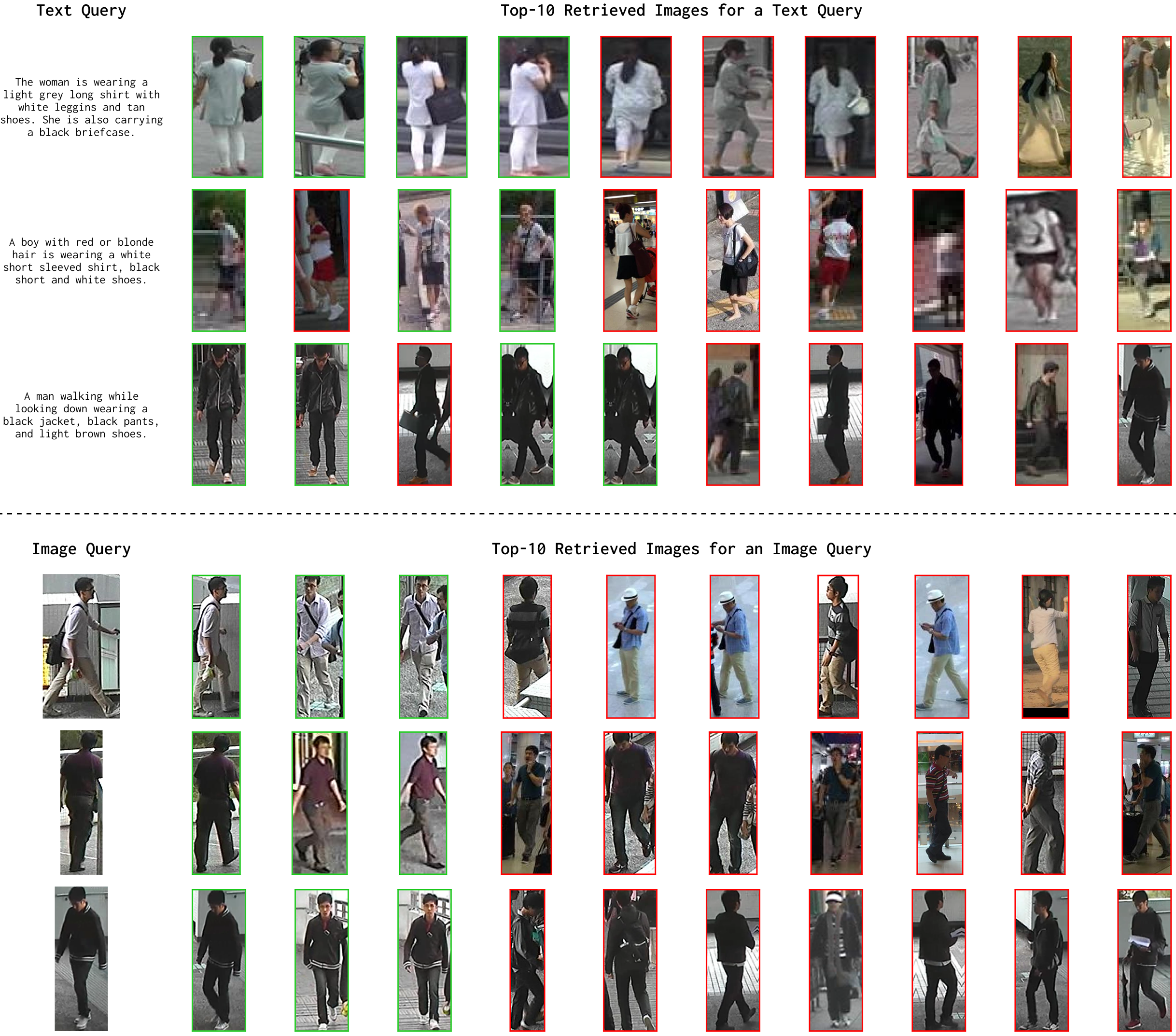}
  \caption{The top-10 retrieved CUHK-PEDES images for text and image queries. The matched and mismatched person images are highlighted with green and red rectangles, respectively.}
  \label{fig:ex}
\end{figure*}

\subsection{Comparison with other I2I and T2I methods} 
\label{sec:sota}
Table~\ref{tab:sota} presents a comparison of meaningful I2I-only, T2I-only, and unified ReID methods across standard benchmarks. Note that UNI is not positioned as a state-of-the-art method. Rather, this comparison aims to show that a model based on a simple idea can achieve competitive performance compared to both I2I- and T2I-only approaches. We include ReText among unified methods because it incorporates a text encoder and can be evaluated for T2I ReID. However, it shows limited effectiveness in T2I retrieval, as it is specifically designed for I2I ReID, which has a fundamentally different objective. Compared to unified methods, the studied method consistently outperforms IRM~\cite{irm} and remains competitive with HPL~\cite{hpl}, demonstrating the ability to perform image- and text-based retrieval. Notably, these results are obtained without introducing additional architectural complexity, but rather through a careful study of the optimization dynamics in unified ReID training. Overall, the results suggest that mitigating conflicts between learning objectives should begin with an analysis of their underlying optimization dynamics, rather than relying solely on architectural modifications. Such an analysis can identify the source of the conflict and motivate simpler, more targeted solutions without unnecessarily increasing model complexity. 
Finally, Figure~\ref{fig:ex} illustrates images from CUHK-PEDES retrieved by UNI for both text and image queries.

\section{Limitations and Future Work}
Our approach relies on freezing the vision encoder during cross-modal alignment. While this effectively preserves the learned visual representation and mitigates the optimization conflict, it limits joint adaptation of the two modalities. As shown in Section~\ref{sec:freeze}, freezing represents one point on the trade-off between I2I and T2I performance, namely the one that fully preserves the visual representation at the cost of not adapting it to the T2I objective. In addition, our study is restricted to person ReID, and the generality of the identified principles to other cross-modal retrieval tasks remains to be investigated. Future work will explore more flexible adaptation strategies that retain the stability of the visual representation while allowing controlled joint optimization, as well as examine whether the observed conflict between instance-level and semantic objectives arises in other open-set cross-modal retrieval problems. Besides, since the textual supervision in the experiments relies on automatically generated captions, the effect of caption quality and the UNI robustness to noisy or corrupted image-text correspondences remain to be investigated.

\section{Conclusion}
In this work, we provided a systematic study of the optimization conflict between I2I and T2I ReID objectives in a unified training framework. This analysis allows us to derive several principles for developing a unified ReID approach. First, separating I2I and T2I optimization through vision encoder freezing during cross-modal training avoids the identified conflict, preserving strong I2I performance while enabling competitive T2I retrieval. Second, sharing domain information between the two stages benefits I2I retrieval, while the scale of text-based data matters more for T2I. Third, incorporating textual supervision during vision encoder pre-training provides a stable semantic anchor for subsequent cross-modal alignment. We assume that these observations can lay the groundwork for a unified multi-modal ReID method. Notably, our findings suggest that competitive performance can be achieved through careful problem study and training strategy, without increasing architectural complexity. We believe that the investigated optimization conflict between instance-level and identity-level objectives reflects a more general issue in information retrieval, where instance-based supervision may be misaligned with semantic relevance in open-set cross-modal retrieval.

\section*{Acknowledgments}
This work was supported by the Ministry of Economic Development of the Russian Federation in accordance with the subsidy agreement~(agreement identifier 000000C313925P4H0002; grant No 139-15-2025-012).

{\small
\bibliographystyle{ieee_fullname}
\bibliography{bibfile}
}

\end{document}